# From Deception to Perception: The Surprising Benefits of Deepfakes for Detecting, Measuring, and Mitigating Bias

Yizhi Liu[1], Balaji Padmanabhan[1], Siva Viswanathan[1*]

[1]Department of Operations & Information Technologies, University of Maryland, College Park; College Park, MD 20742, USA.

**Abstract:** While deepfake technologies have predominantly been criticized for potential misuse, our study demonstrates their significant potential as tools for detecting, measuring, and mitigating biases in key societal domains. By employing deepfake technology to generate controlled facial images, we extend the scope of traditional correspondence studies beyond mere textual manipulations. This enhancement is crucial in scenarios such as pain assessments, where subjective biases triggered by sensitive features in facial images can profoundly affect outcomes. Our results reveal that deepfakes not only maintain the effectiveness of correspondence studies but also introduce groundbreaking advancements in bias measurement and correction techniques. This study emphasizes the constructive role of deepfake technologies as essential tools for advancing societal equity and fairness.

## 1. Introduction

Individuals from minority groups, even with equivalent qualifications, consistently receive fewer opportunities in critical areas such as employment, education, and healthcare. Yet, empirically demonstrating the existence of such pervasive bias, let alone measuring the extent of bias or correcting it, remains a significant challenge. Over several decades, researchers have utilized a range of experimental methodologies to test for biases in real-life situations (Bertrand and Duflo 2017). Audit studies, among the earliest of such methods, match two individuals who are similar in all respects except for sensitive characteristics like race, to test decision-makers' biases (Ayres and Siegelman 1995). A significant limitation of this method, however, is the inherent impossibility of achieving an exact match between two individuals, precluding perfect comparability (Heckman 1998).

Correspondence studies have emerged as a predominant experimental approach for measuring biases (Guryan and Charles 2013, Bertrand and Mullainathan 2004). They create identical fictional profiles with manipulated attributes like race to assess differential treatment. However, these studies traditionally manipulate solely textual information, which may not reflect contemporary decision-making scenarios increasingly influenced by visual cues like facial images, as seen in recent hiring processes (Acquisti and Fong 2020, Ruffle and Shtudiner 2015). This reliance on text limits their effectiveness, as modern contexts often involve multimedia elements, making it challenging to measure real-world biases accurately or correct them based on such incomplete information (Armbruster et al. 2015). Nevertheless, incorporating images into correspondence studies presents significant challenges due to the difficulty of creating consistently comparable manipulat[1]ed images. Despite several efforts in this direction, such as modifying or controlling

---

[*] Corresponding Author: Siva Viswanathan. Contact: sviswan1@umd.edu

features like attractiveness (Mejia and Parker 2021, Rooth 2009), these approaches have yet to meet the high standards of controlled comparability required for correspondence studies.

With advancements in artificial intelligence, particularly through deepfake technology known for generating realistic facial images, the opportunity to create highly comparable images is now within reach. While deepfake technology is often surrounded by privacy and security concerns, its potential for societal good is underexplored (Zhao et al. 2021, Westerlund 2019). We harness deepfake technology to modify facial images in a controlled manner, allowing us to measure bias more accurately in correspondence studies that include such visual information. This approach allows researchers to detect, measure, and potentially correct biases in contexts where facial images play a critical role, thereby advancing the methodology of correspondence studies to better align with contemporary societal interactions.

In this study, we explore how deepfake technology can be applied to measure bias in pain assessment, where assessors rely on facial images of patients to evaluate pain levels. Pain assessment's inherent subjectivity necessitates meticulous control for bias measurement, aligning with the idea of correspondence studies (Giordano et al. 2010). For example, if assessors assign different pain levels to facial images of two patients—one Black and one White—it's unclear whether these differences are due to actual pain levels or assessor bias. Therefore, high comparability between each pair of images is crucial, yet challenging to achieve. Deepfake technology can potentially facilitate this by creating highly consistent and controlled facial images. By conducting experiments on crowdsourcing platforms, we demonstrate that deepfakes can be effectively used to isolate and measure bias in pain assessments. More importantly, our research advances the methodology of correspondence studies by enabling more accurate measurement of bias in scenarios that include image information, thereby making it more applicable in modern settings, paving the way for future research. Additionally, our study demonstrates how a technology predominantly perceived for its risks can play a pivotal role in enhancing fairness and reducing discrimination within critical societal sectors.

## 2. Literature Review

### 2.1 Bias Measurement and Correction

While bias is widely acknowledged as pervasive, its precise measurement and correction in real-world contexts remain methodologically challenging. Correspondence studies have emerged as the current gold standard for empirically examining bias, enabling researchers to isolate the causal effects of sensitive attributes (e.g., race) on decision-making outcomes (Verhaeghe 2022). These studies operationalize bias measurement by constructing pairs of fictitious profiles that are identical in all respects except for the manipulated sensitive attribute, such as race (Bertrand and Duflo 2017). For example, a seminal work by Bertrand and Mullainathan (2004) demonstrates racial bias in hiring by altering racially coded names on resumes while holding other qualifications constant. However, traditional correspondence studies face a critical limitation: their reliance on textual manipulations (e.g., names) may not account for modern decision-making environments where visual cues, such as facial images, increasingly influence outcomes. For example, in important areas like healthcare, employment, and social media, decision-makers often evaluate individuals through multimedia profiles, where facial features can shape judgments (Acquisti and Fong 2020, Armbruster and Delage 2015). This gap limits the validity of text-only correspondence studies, as biases triggered by visual cues remain unaddressed.

Recent advances in correspondence studies have attempted to address this limitation by incorporating visual cues in experimental designs. For example, Busetta et al. (2021) and Baert (2018) investigate how facial attractiveness in profile pictures affects hiring biases, while the other features, such as race and gender, are controlled for comparability. Polavieja et al. (2023) examined racial bias triggered by resume photos, which are also matched on attractiveness, competence, and sympathy. In Bellemare et al. (2023), professional actors are randomly assigned a standard chair or wheelchair to measure disability bias in hiring. However, the issues of visual cues remain, since individuals, even when matched for similarity, have inherent variations in facial features and expressions that may confound bias measurement.

The bias research in management disciplines has increasingly emphasized the role of visual cues in modern interactive decision-making (Malik et al. 2023, Gunarathne et al. 2022). Researchers have employed diverse strategies to integrate visual elements while attempting to isolate specific biases. For instance, Mejia and Parker (2021) utilize profile photos standardized for attractiveness by varying race and gender to examine racial and gender bias in ridesharing platforms. Similarly, Acquisti and Fong (2020) obtain profile photographs of models with median levels of attractiveness and professionalism to test hiring biases in online labor markets. Recognizing the importance of visual cues, some studies deliberately exclude human faces from profile images in their investigations of racial bias on Airbnb (Cui et al. 2020, Edelman et al. 2017). Nevertheless, these prior studies remain significantly limited in their capability to rigorously control comparability. When visual stimuli are not meticulously adjusted to vary solely by the targeted attribute, the results might inadvertently blend true bias with other variables such as differences in facial expressions or lighting conditions. Such meticulous controls are particularly important in areas where subtle visual cues play a critical role, such as in pain assessment. In such contexts, precisely isolating and measuring bias is crucial for avoiding misinterpretations and ensuring accurate decision-making, highlighting the need for more refined methodologies in bias research.

## 2.2 Pain Assessment

Pain assessment is a longstanding task in healthcare. An accurate assessment of pain plays a pivotal role in diagnosing conditions, guiding treatment decisions, and monitoring patient recovery (Breivik et al. 2008). Despite its importance, assessing pain objectively is a non-trivial task because pain is a subjective experience (American Pain Society 1999). The most direct and reliable method for pain assessment is a patient's self-report, but unfortunately, patients are not always able to describe their pain (Herr et al. 2011). For example, patients with limited communication abilities, such as infants and patients with cognitive impairment, may not be able to verbalize their pain properly (Severgnini et al. 2016). To address this challenge, observational scales have been proposed as alternatives to self-reporting. These scales, such as the Behavioral Pain Scale and Children's Revised Impact of Event Scale, require well-trained observers, usually clinical experts, to measure patients' pain levels through their behaviors (Perrin et al. 2005, Payen et al. 2001). This approach is still limited by the expertise and availability of the observers, as observers may not always be available, especially for in-home patients, restricting timely assessments (Fritz et al. 2020).

In recent years, AI-based pain assessment has been proposed as an automated solution to pain assessment that can facilitate early clinical intervention (Werner et al. 2019). It usually uses algorithms to analyze patients' pain levels in facial images, offering standardized pain scoring, and it is available anytime, anywhere, addressing the limitations of previous manual approaches (De Sario et al. 2023). Specifically, most AI pain assessment models are computer vision-based models that learn and extract behavioral features of pain from patients' facial images, usually manifested in their facial expressions (Huo et al. 2024). For instance, early studies use support vector machine (SVM) classifiers to determine the patient's pain level in facial images, where the statistics of action units (AUs) or active appearance models (AAM), which locate facial landmarks, are used as facial features (Lucey et al. 2010, Littlewort 2009). More recently, with the development of deep learning technology, models such as convolutional neural networks (CNN) can automatically extract features and complete computer vision tasks in an end-to-end manner. Researchers have gradually adopted this method to predict pain and achieved better results (Fontaine et al. 2022, Rodriguez et al. 2022, Semwal and Londhe 2021, Bargshady et al. 2020).

Nonetheless, accurately measuring bias in pain assessment poses unique challenges. Both human assessors and AI models rely heavily on subtle facial features, such as the activation of specific facial muscles (i.e., AUs), to estimate pain intensity (Ekman and Friesen 1978). These features can vary significantly across individuals, even when images appear superficially similar in attributes like pain intensity and attractiveness (Chu et al. 2013). Therefore, if researchers compare such images, observed differences in pain intensity could reflect these inherent variations rather than true bias related to a target attribute. To isolate bias, a desired approach should ensure that compared images differ only in the sensitive attribute of interest (e.g., skin tone), while maintaining other attributes identical. Otherwise, it becomes impossible to distinguish whether disparities in pain assessments stem from bias or confounding factors.

## 3. Methodology

### 3.1 Data

In collaboration with a digital pain management company, we obtain 100 facial images showing painful expressions to facilitate our study. To ensure the controllability of our research conditions, the subjects depicted in these images are exclusively of white and black races, with no representation from other ones. In addition, since age is a continuous variable, to explore age bias, we divide these images into two categories - young and senior (old). We use VGGFace to verify the age of each image according to the definition of the US census[2], classifying our original images into the young or senior category (Cao et al. 2018). Specifically, the age categories are: 18-24, 25-34, 34-44, 45-54, 55-64, and 65 or above. We define young as 18-34 and senior as 55 and above, and all of our images belong to one of these two categories. The dataset also represents both male and female subjects, integrating gender considerations into our research. We unify all images to ensure the uniformity of their resolution and size. Then, to eliminate the influence of image information other than faces, we uniformly use grayscale processing on the pictures.

### 3.2 Manipulation Method

Using Generative Adversarial Networks (GAN)-based deepfake techniques, we generate three derivatives for each image: those altered for race, age, and a combination of both (Karras et al. 2020). Our manipulation purpose is to alter the assessors' perceptions of the subjects' race and age, while the pain levels of the subjects should remain consistent, which allows us to measure bias at the individual intensity through experiments. It is worth noting that although we can directly manipulate the race of a facial image, this may cause changes in facial features and thus affect the pain intensity. Therefore, we choose to manipulate skin tone instead.

---

[2] Source: https://www.census.gov/newsroom/blogs/random-samplings/2023/09/exploring-diversity.html, accessed on Oct 20, 2024.

To this end, we utilize facial editing algorithms, which have become a popular research topic in recent years, and GAN-based generative AI such as StyleGAN has been widely used for this purpose (Abdal et al. 2019). Notably, these models rely on image inversion, which involves mapping the image back to the model's latent space for editing (Dinh et al. 2022). When the image to be edited is very different from the training data, image inversion may be inaccurate, resulting in unnaturally generated images. To address this problem, we use StyleFeatureEditor, a state-of-the-art solution that simultaneously achieves both excellent reconstruction quality and good editability (Bobkov et al. 2024). This allows us to create high-quality adjustments to real faces. Note that StyleFeatureEditor is a training pipeline for StyleGAN-based models, and it still relies on several underlying models for face editing, such as E4e (Tov et al. 2021), StyleCLIP (Patashnik et al. 2021), and InterfaceGAN (Shen et al. 2020). Specifically, we use StyleFeatureEditor with E4e to manipulate the hair color, StyleCLIP to manipulate the hairstyle or skin tone, and InterfaceGAN to manipulate the age. We present one example for each type of manipulation in Figure 1. We carefully manipulate only the skin tone to change racial perceptions, avoiding alterations to facial features that might influence perceived pain levels. Also, to minimize racial cues from hairstyles, we adopt more neutral hairstyles in the manipulated conditions. When manipulating age, we check to ensure that the adjusted age is in the corresponding range. For example, if the original age of the image is 18-34 years old, we define it as young and ensure that it is in the senior range after manipulation, that is, 55 years old or above. Consequently, we expand each original image to four distinct variations, as detailed in Figure 2, creating a total of 400 images.

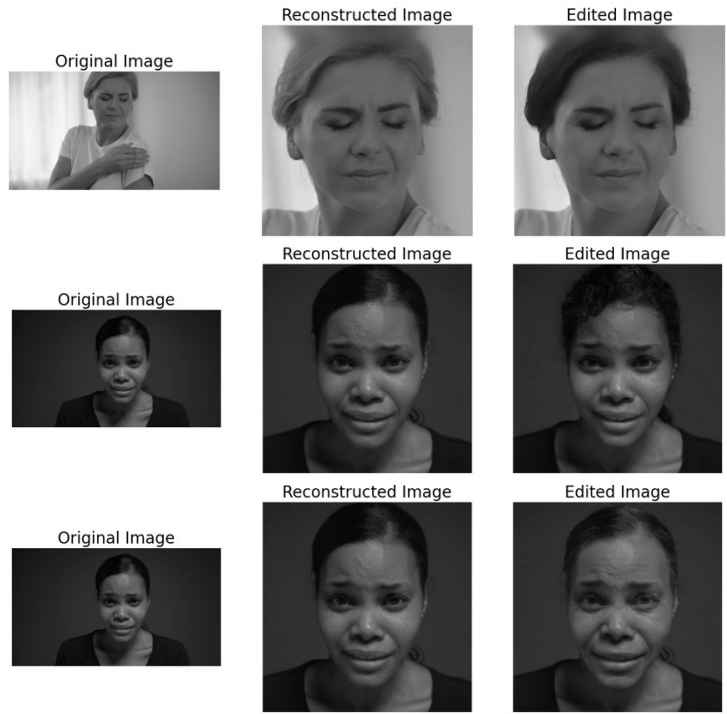

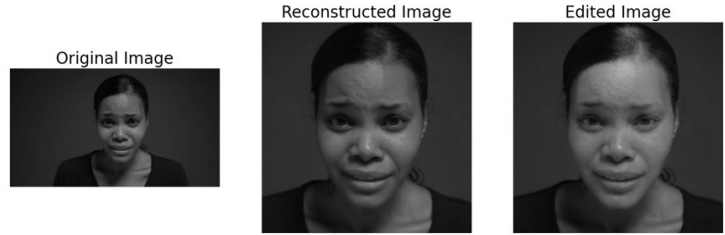

**Fig. 1. Examples of image manipulations.** First row: hair color manipulated; second row: hair style manipulated; third row: age manipulated; fourth row: skin tone manipulated.

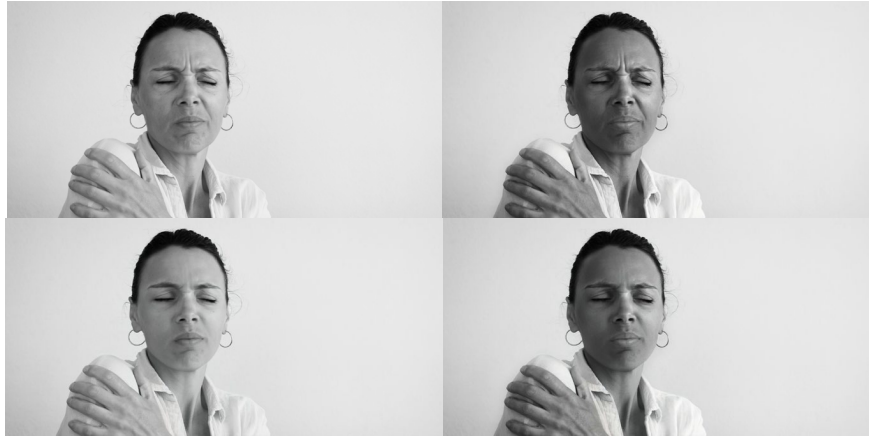

**Fig. 2. Examples of image manipulations.** Upper left: original; upper right: skin tone manipulated; lower left: age manipulated; lower right: skin tone and age manipulated.

### 3.3 Experimental Design

After data collection and manipulation using deepfake technology, each image is randomly assigned to an independent assessor to assess the pain level. This assessment focuses on analyzing facial Action Units (AUs), based on the Facial Action Coding System (Ekman and Friesen 1978), developed to describe any facial expression in terms of anatomically derived action units. These AUs, as it turns out, are crucial for calculating the Prkachin and Solomon Pain Intensity (PSPI) score - a recognized measure of pain intensity (Hammal and Cohn 2012). Figure 3 illustrates which parts of a face different AUs represent. PSPI is calculated by AU4 + MAX(AU6, AU7) + MAX(AU9, AU10) + AU43, where, AU43 is either 0 or 1, and the other AUs are a categorical number ranging from 0 to 5.

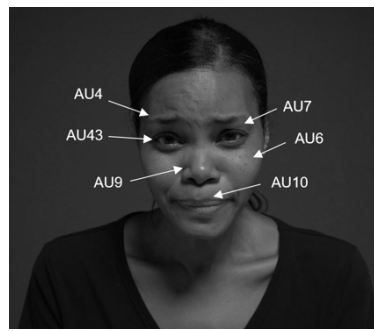

**Fig. 3. Illustration of action units (AU)**

We conduct the assessment tasks on Amazon Mechanical Turk (AMT) and Credamo. Both these platforms have been extensively used in previous studies (Douglas et al. 2023, Tang et al. 2023). In order to ensure high-quality assessments, we only select assessors with a historical pass rate above 95%. A screening test and a manipulation test precede the formal assessments to confirm the assessors' comprehension of the tasks (we exclude data from those who failed these tests). Considering the diverse racial backgrounds of Turkers, we anticipate that measured biases could be influenced by this variability. To mitigate this, we also conduct identical experiments on Credamo, a platform similar to AMT, except in this case all the assessors are Chinese (Jin et al. 2020). Figure 4 illustrates an example screenshot of our task.

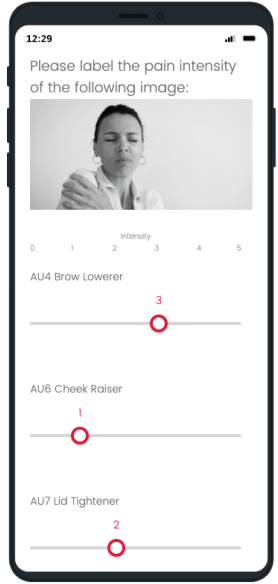

**Fig. 4. An example screenshot of our task**

## 4. Results

### 4.1 Evaluation of Manipulation Quality

We recruited 2,000 assessors on AMT and 1,200 on Credamo. On average, each image is evaluated by five assessors on AMT and three on Credamo. Prior to analysis, we first evaluate the authenticity of our edited images to ensure they are not perceived as fakes, which could compromise our results. To this end, we employ a state-of-the-art Vision Transformer (ViT) model for deepfake detection (Heo et al. 2023) that boasts a 97.8% F1 score on the Celeb-DF dataset, a standard benchmark for deepfake detection (Li et al. 2023). Figure 5 provides an example of the deepfake detection results. At the end of the pain assessment task, we also ask the assessors about the image quality. All 300 of our manipulation images pass these rigorous tests and are not identified as fakes by the model or assessors.

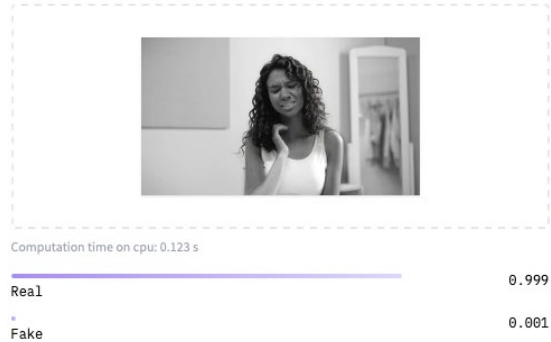

**Fig. 5. An example of deepfake detection results**

We further utilize face verification of VGGFace to test whether a pair of facial images (i.e., before vs after manipulation) are considered the same person by AI models. Our results show that AI indeed considers them as the same person. This is similar to what humans do – when a pair of such images are presented to humans, we know they are the same person as well. It also ensures that our manipulations are realistic sufficiently. Moreover, we use BlazeFace of Google AI Edge to test whether our manipulations have changed the facial landmarks of each pair of images (Bazarevsky et al. 2019). Figure 6 illustrates an example output of facial landmarks. For each pair of facial images (i.e., before vs after manipulation), we calculate the average Euclidian distance of the facial landmarks to measure the difference between them. If our manipulation is successful, we should not observe any significant difference in facial landmarks, which could lead to altered pain levels. As shown in Figure 7, there is no significant difference in facial landmarks between different conditions, which indicates that our manipulation itself did not significantly change the pain levels.

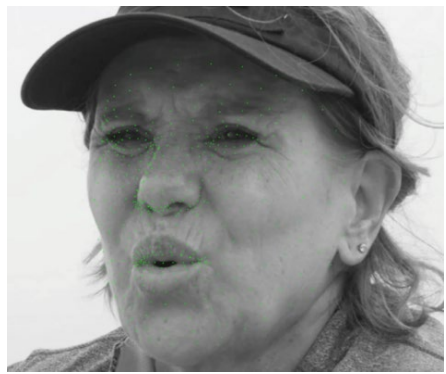

**Fig. 6. An example of facial landmarks from BlazeFace**

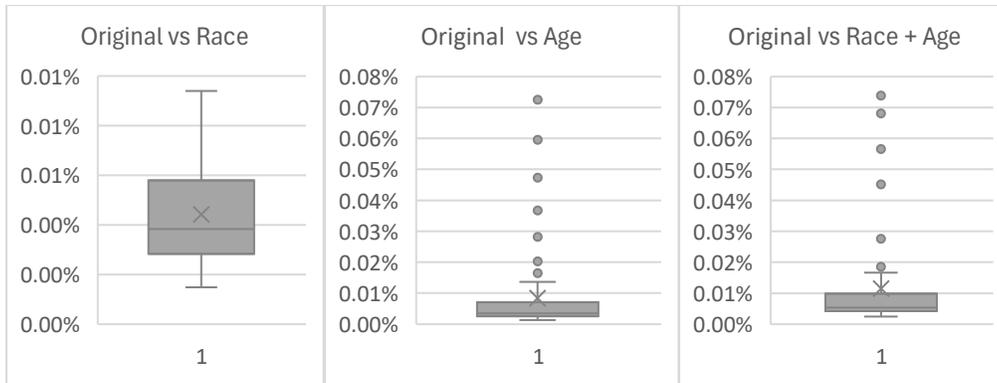

**Fig. 7. Comparison of facial landmarks between different conditions**

## 4.2 Measurement of Bias

To assess racial bias, we compare the average PSPI scores between black and white subjects. The results, as shown in Table 1, reveal that PSPI scores for white subjects are consistently higher—by 10.09% on AMT (1.093/10.833) and 10.90% on Credamo (1.042/9.562) than those for black subjects. This consistent disparity across both platforms underscores the presence of racial bias in pain assessment, with a notably greater bias observed for male subjects compared to females, particularly on Credamo.

**Table 1. Measurement of racial bias.** * $p < 0.1$; ** $p < 0.05$; *** $p < 0.01$.

| Sample | PSPI for Black | PSPI for White | Difference |
|---|---|---|---|
| **Panel A. AMT** | | | |
| All Subjects | 10.833 [1,000] | 11.926 [1,000] | -1.093*** (0.116) |
| Female | 10.786 [660] | 11.876 [660] | -1.089*** (0.145) |
| Male | 10.923 [340] | 12.023 [340] | -1.100*** (0.193) |
| **Panel B. Credamo** | | | |
| All Subjects | 9.562 [600] | 10.603 [600] | -1.042*** (0.159) |
| Female | 9.654 [396] | 10.604 [396] | -0.949*** (0.201) |
| Male | 9.382 [204] | 10.603 [204] | -1.221*** (0.259) |

Our analysis also extends to potential age bias. As presented in Table 2, subjects perceived as senior (i.e., older) received higher pain assessment scores—3.33% on AMT (0.373/11.193) and 5.24% (0.515/9.825) on Credamo—compared to when they are perceived as younger. Although the magnitude of age bias is less pronounced than racial bias, it is consistently observed, particularly on Credamo, suggesting potential cultural influences on assessors' perceptions. Similar to the patterns of racial bias, this age bias is more pronounced among male subjects.

**Table 2. Measurement of age bias.** * $p < 0.1$; ** $p < 0.05$; *** $p < 0.01$.

| Sample | PSPI for Senior | PSPI for Young | Difference |
|---|---|---|---|
| | **Panel A. AMT** | | |
| All Subjects | 11.566 [1,000] | 11.193 [1,000] | 0.373*** (0.118) |
| Female | 11.495 [660] | 11.167 [660] | 0.329** (0.148) |
| Male | 11.703 [340] | 11.244 [340] | 0.459** (0.197) |
| | **Panel B. Credamo** | | |
| All Subjects | 10.340 [600] | 9.825 [600] | 0.515*** (0.161) |
| Female | 10.369 [396] | 9.889 [396] | 0.480** (0.203) |
| Male | 10.284 [204] | 9.701 [204] | 0.583** (0.264) |

To gain a more complete understanding of bias in pain assessment, especially the cross-effects of race and age, we examined PSPI Ranking by Subject Race and Age. As can be seen in Table 3, senior white subjects were always given the highest scores, followed closely by young white subjects. After these groups were the senior black subjects and young black subjects. This ranking order was largely consistent across both AMT and Credamo platforms, as well as across male and female subsets.

**Table 3. PSPI Ranking by Subject Race and Age.** W stands for white and B stands for black. Y and S represent young and senior, respectively.

| Rank | US (AMT) | | China (Credamo) | |
|---|---|---|---|---|
| | Group | Mean PSPI | Group | Mean PSPI |
| | **Panel A: All Samples** | | | |
| 1 | WO | 12.04 [500] | WO | 10.89 [300] |
| 2 | WY | 11.81 [500] | WY | 10.32 [300] |
| 3 | BO | 11.09 [500] | BO | 9.79 [300] |
| 4 | BY | 10.58 [500] | BY | 9.33 [300] |
| | **Panel B: Female Subsamples** | | | |
| 1 | WO | 11.99 [330] | WO | 10.83 [198] |
| 2 | WY | 11.76 [330] | WY | 10.37 [198] |
| 3 | BO | 11.00 [330] | BO | 9.90 [198] |
| 4 | BY | 10.57 [330] | BY | 9.40 [198] |
| | **Panel C: Male Subsamples** | | | |

| | | | | | |
|---|---|---|---|---|---|
| 1 | WO | 12.15 [170] | WO | 10.99 [102] |
| 2 | WY | 11.90 [170] | WY | 10.22 [102] |
| 3 | BO | 11.26 [170] | BO | 9.58 [102] |
| 4 | BY | 10.59 [170] | BY | 9.19 [102] |

Our findings of racial bias in pain assessment are consistent with previous qualitative findings (Hoffman et al. 2016). Unlike racial bias, there is less empirical evidence of age bias in pain assessment, while our study also effectively discovered and quantified this age bias. Our methodology that leverages deepfakes not only facilitates the identification and quantification of these biases, but, as we show further below, also provides actionable insights for medical practitioners and policymakers to address and mitigate such biases.

### 4.3 Bias Measurement at the AU Level

Our experimental design enables assessors to participate fully in the evaluation process, allowing for an in-depth analysis at the AU level. This analysis helps understand the underlying mechanisms of bias in pain assessment. For instance, as shown in Table 4, white subjects typically receive higher scores on AUs such as the Brow Lowerer (AU4), Lid Tightener (AU7), and Nose Wrinkler (AU9). Conversely, black subjects are scored higher on the Cheek Raiser (AU6) and Upper Lip Raiser (AU10). For senior subjects, although overall scores are higher across all AUs, significant differences are particularly notable in AU4, AU7, and AU9. Although previous studies have found that people may exhibit bias toward AUs, our study provides empirical evidence of such bias in pain assessment (Chen and Joo 2021), an important application with meaningful differences in how subjects may subsequently receive treatment. Moreover, the detailed exploration of AUs also provides a granular view of the sources of bias in pain assessment.

**Table 4. Measurement of racial and age bias at the AU level.** W stands for white, B stands for black. Y and S represent young and senior, respectively.

| | AU4 | AU6 | AU7 | AU9 | AU10 | AU43 |
|---|---|---|---|---|---|---|
| **Panel A. AMT** | | | | | | |
| t stat (W v B) | 7.308 | -1.495 | 6.379 | 10.360 | -2.246 | 0.940 |
| p Values | (0.025) | (0.068) | (0.000) | (0.000) | (0.001) | (0.174) |
| t stat (Y v S) | -1.890 | -1.199 | -1.440 | -0.564 | -2.930 | -0.492 |
| p Values | (0.029) | (0.115) | (0.075) | (0.286) | (0.002) | (0.311) |
| **Panel B. Credamo** | | | | | | |
| t stat (W v B) | 5.974 | -1.934 | 7.133 | 12.133 | -1.811 | 1.628 |
| p Values | (0.000) | (0.027) | (0.000) | (0.000) | (0.035) | (0.052) |
| t stat (Y v S) | -3.281 | -1.494 | -0.707 | -5.733 | -1.448 | -0.929 |
| p Values | (0.001) | (0.068) | (0.240) | (0.000) | (0.074) | (0.176) |

### 4.4 Bias Correction and AI Fairness

After assessing bias in our pain prediction models, we turn to explore if such bias can be corrected. Specifically, we use the assessment results as labeled data to train and test AI models for pain assessment. This effort stems from two main factors. First, training AI models for automated pain assessment is a key objective in pain assessment from facial images. This automation could enable accurate pain assessments where patients cannot verbalize their pain or where there is a shortage of medical expert resources (Zhang et al. 2023). Second, employing these models allows for the immediate observation of bias correction outcomes. Although human biases can also be corrected through education, such processes typically do not yield instant results.

We utilize the ResNet50 model, noted for its efficacy in pain assessment studies (Gkikas and Tsiknakis 2023). Our training and evaluation dataset comprises 90% of our total data, including 360 images. The remaining 10%, or 40 images, form our testing set. Our experiments include four different conditions: Original, Average, Autocorrection, and a combination of Average and Autocorrection. To prevent the model from being trained on multiple labels for the same subject, which could introduce additional bias, we use only one label for each image.

Specifically, for the original condition, the label is the average of the PSPI scores that an image received from the AMT assessments, a common approach similar to majority voting intended to minimize bias (Davani et al. 2022). Central to our experiment is the evaluation of whether averaging the labels obtained from black and white subjects could reduce bias in the model's pain assessments, where we focus on the potential correction of racial bias. To this end, we introduce the Average condition, where the PSPI scores from the image and its race-manipulated counterpart are averaged to serve as labels for both images. Additionally, inspired by previous research, we implement an Autocorrection condition (Fu et al. 2020). This approach involves incorporating sensitive features (e.g., race) directly into the model and assigning learnable weights to them, anticipating that the model will adjust itself to mitigate bias. Finally, we combine the Average and Autocorrection to observe their cumulative effect.

An example of how these labels are computed is shown in Table 5. Suppose we have a facial image with its pain intensity in the white condition (original) is 14, and the pain score in the black condition is 12. We use 14 as the label for the original condition, and their average 13 as the label for the average condition. For conditions 3 and 4, we add additional information such as the subject's race, gender, and age as features to the model. These features will be coded and given different weights during the model learning process through learnable weights. A pseudo-code of our bias correction process is presented in Table 6. Note that these methods only apply to the training set. To ensure comparability across conditions, we engaged a pain management doctor in China to assess the pain depicted in the testing set images. These assessments serve as consistent labels across different conditions.

**Table 5. A bias correction example**

| Set ID | Method | Used Image(s) | Example Label |
|---|---|---|---|
| 1 | Original | Original | 14 |
| 2 | Average | Original, race manipulated | (14+12)/2 = 13 |
| 3 | Auto | Original | 14; Race: White, Gender: Female, Age: Young |

| 4 | Average + Auto | Original, race manipulated | (14+12)/2 = 13; Race: White, Gender: Female, Age: Young |

**Table 6. Pseudo-code of bias correction**

| Algorithm 1: Bias Correction |
|---|
| 1: **Input**: Image sets I, S, A, SA; label sets PI, PS, PA, PSA; model *pm* |
| 2: **Output**: MSE, RMSE, and Individual Fairness metrics |
| 3: Randomly select 10% of images from I, define as I'. |
| 4: **for** each image $i_j \in$ I': |
| 5:     Identify $s_j$ from S, $a_j$ from A, $s_{aj}$ from SA to form the testing set T |
| 6:     Define the remaining 80% images in I, S, A, SA as the training set R. |
| 7: **end for** |
| 8: // Original condition |
| 9: Train *pm* on R using labels from Ytrue = {PI, PS, PA, PSA} |
| 10: Predict on T, get Ypred = {PredI, PredS, PredA, PredSA} |
| 11: Calculate: |
| 12:     $MSE_{org} = \frac{1}{|T|} * \sum(Ytrue_j - Ypred_j)^2 \ for\ all\ i \in T$ |
| 13:     $RMSE_{org} = sqrt(MSE_{org})$ |
| 14: |
| 15:     $IF_{org} = \frac{1}{|T|} * \sum |PredI_j - PredS_j| + |PredA_j - PredSA_j| \ for\ all\ i \in T$ |
| 16: // Average condition |
| 17: // Take the average labels over race |
| 18: **for** each image index j in PI, PS, PA, PSA: |
| 19:     $AvgPIPS_j = (PI_j + PS_j) / 2$ |
| 20:     $AvgPAPSA_j = (PA_j + PSA_j) / 2$ |
| 21:     // Update labels |
| 22:     $PI_j = AvgPIPS_j$ |
| 23:     $PS_j = AvgPIPS_j$ |
| 24:     $PA_j = AvgPAPSA_j$ |
| 25:     $PSA_j = AvgPAPSA_j$ |
| 26: **end for** |
| 27: Replicate the training and testing procedure; skip for simplicity |
| 28: |
| 29: // Autocorrection condition |
| 30: Extract $F_j$ = {race, age, gender} for each image in $i_j$ R and T |
| 31: Train pm using $F_j$ and Ytrue = {PI, PS, PA, PSA} |
| 32: Replicate the testing procedure; skip for simplicity |
| 33: |
| 34: // Average + Autocorrection condition |
| 35: Extract $F_j$ = {race, age, gender} for each image in $i_j$ R and T |
| 36: // Take the average labels over race |
| 37: **for** each image index j in PI, PS, PA, PSA: |
| 38:     $AvgPIPS_j = (PI_j + PS_j) / 2$ |
| 39:     $AvgPAPSA_j = (PA_j + PSA_j) / 2$ |
| 40:     // Update labels |
| 41:     $PI_j = AvgPIPS_j$ |
| 42:     $PS_j = AvgPIPS_j$ |
| 43:     $PA_j = AvgPAPSA_j$ |
| 44:     $PSA_j = AvgPAPSA_j$ |
| 45: **end for** |

| 46: | Train *pm* using F$_j$ and Ytrue = {PI, PS, PA, PSA} |
| 47: | Replicate the testing procedure; skip for simplicity |

Since the PSPI score is a continuous value ranging from 0 to 16, we utilize mean squared error (MSE) and root mean squared error (RMSE) to evaluate our model's performance. However, as we cannot guarantee that the labels for the testing sets closely approximate the ground truth (Lebovitz et al. 2021), our primary metric of evaluation is the fairness of the model. Using our deepfake methodology, we can in fact assess fairness for each individual. In particular, we evaluate individual fairness by comparing the predicted pain levels for an image against those for its race-manipulated counterpart, with the absolute difference between these predictions serving as a measure of individual fairness. This approach, focusing on individual rather than group-based fairness, has been widely acknowledged for its benefits (Li et al. 2023, Fu et al. 2020, Dwork et al. 2012).

The results, presented in Table 7, indicate that the average method significantly improved individual fairness by 32.96% ((1.291%-0.971%)/0.971%). Interestingly, the autocorrection method alone could improve model performance but tremendously exacerbate the model's bias. However, when combined with the average method, the model tends to achieve a balance between performance and fairness. These findings affirm that our deepfake-based method for measuring bias can not only assess but also be used to correct bias. Moreover, such benefits can extend from human assessments to AI models through the training process.

**Table 7. The performance of AI pain assessment.** Model settings and hyperparameters are the same across all conditions. Five-fold cross-validation is applied to obtain robust results.

| Set ID | Condition | Training | | Testing | | |
|---|---|---|---|---|---|---|
| | | MSE | RMSE | MSE | RMSE | Individual Fairness |
| 1 | Original | 4.627 | 2.062 | 13.031 | 3.610 | **1.291%** |
| 2 | Average | 1.697 | 1.246 | 11.647 | 3.413 | **0.971%** |
| 3 | Auto | 3.506 | 1.810 | 7.858 | 2.803 | **13.449%** |
| 4 | Average + Auto | 1.223 | 1.098 | 7.815 | 2.795 | **3.057%** |

# 5. Conclusion and Discussion

Sound methodologies for empirically testing for the existence of bias are crucial for society. Over the past several decades, the emergence of field experimental methods such as correspondence studies has significantly advanced this agenda. These studies have been widely applied in critical areas such as employment, education, and healthcare, often proving the existence of biases and substantially enhancing our understanding of discrimination in our society. However, correspondence studies traditionally rely on modifying textual information, such as the names of fictitious applicants, to ensure comparability. With the evolution of the internet and the growing importance of visual information, this text-based limitation is increasingly unable to meet the demands of many real-world scenarios, thus restricting the applicability of correspondence studies and the measurement and correction of actual biases. Despite some recent attempts, incorporating visual cues into experimental bias measurement remains a significant challenge due to the difficulty of creating highly comparable images.

In this research, we demonstrate how deepfake technology, often viewed through a lens of criticism due to its association with unethical and illegal activities, can be a powerful tool for social good when applied thoughtfully and responsibly (Wiens et al. 2019). In the critical context of pain assessment, we used deepfakes to construct facial images under varying race and age conditions. By utilizing these realistic image pairs, we conducted experiments on crowdsourcing platforms. The results indicate that our method can effectively detect and even quantify specific racial and age biases, which are unachievable through methods used in previous correspondence studies. Moreover, when we used these human decisions to train and test AI models for pain assessment, we found that by averaging the labels of the image and its race-manipulated counterpart, we could effectively improve individual fairness. This offers a new approach to bias correction research.

Our proposed methodology extends beyond pain assessment in the healthcare sector, offering immense potential to revolutionize bias measurement across various critical domains, such as criminal justice, education, and employment. In these fields, where biases can lead to life-changing consequences, ensuring fairness is imperative. For example, in the criminal justice system, the unbiased assessment of facial images could prevent miscarriages of justice by mitigating undue influence on judicial decisions (Linna et al. 2024, Rhode 2008). Similarly, in education and employment, our approach could critically evaluate biases in the admissions and hiring processes, where the visual representation of candidates often plays an important role (Ong 2022, Ruffle and Shtudiner 2015, Stevenage and McKay 1999). In addition, while this paper primarily focuses on image manipulation, the underlying principles of our methodology are readily adaptable to video content. Applying our approach to videos, however, involves more sophisticated manipulations and the integration of advanced deepfake technologies. We are exploring these enhancements with a view to broadening the applicability of our research, thereby amplifying its real-world impact. Ultimately, this study may set a foundation for future studies aimed at transforming the measurement and correction of biases, with the potential to significantly influence public perception and decision-making processes. Our research also underscores the importance of adopting an open and innovative approach toward technologies that may initially seem adverse. By ethically leveraging technologies such as deepfakes, we can harness their potential to significantly benefit society (Xie and Avila 2024).